\documentclass[11pt,a4paper]{article}
\usepackage[hyperref]{emnlp2018}
\usepackage{amsmath}
\usepackage{amssymb}
\usepackage{times}
\usepackage{relsize}
\usepackage{latexsym}
\usepackage{bbm}
\usepackage{graphicx}
\usepackage{booktabs}
\usepackage{multirow}
\usepackage{esvect}
\usepackage{url}

\newcommand{\ti}[1] {\textit{#1}}
\newcommand{\tm}[1] {\texttt{#1}}

\newcommand{\MC}[3]{\multicolumn{#1}{#2}{#3}}
\newcommand{\MR}[3]{\multirow{#1}{#2}{#3}}

\newcommand{\ra}{$\rightarrow$}

\aclfinalcopy 

\title{LIUM-CVC Submissions for WMT18 Multimodal Translation Task}

\author{Ozan Caglayan, Adrien Bardet, \\
         \bf Fethi Bougares, Lo\"ic Barrault \\
         LIUM, Le Mans University \\
         \tm{FirstName.LastName@univ-lemans.fr} \\
          \\
         \bf Kai Wang, Marc Masana, Luis Herranz and Joost van de Weijer\\
    CVC, Universitat Autonoma de Barcelona\\
    {\tt \{kwang,mmasana,lherranz,joost\}@cvc.uab.es}}

\date{}

\begin{document}
\maketitle

\begin{abstract}
This paper describes the multimodal Neural Machine Translation systems
developed by LIUM and CVC for WMT18 Shared Task on Multimodal Translation.
This year we propose several modifications to our previous multimodal attention
architecture in order to better integrate convolutional features and
refine them using encoder-side information.
Our final constrained submissions ranked first for English\ra French and second for English\ra German
language pairs among the constrained submissions according to the automatic evaluation metric METEOR.
\end{abstract}

\section{Introduction}
In this paper, we present the neural machine translation (NMT) and multimodal NMT (MMT) systems
developed by LIUM and CVC for the third edition of the shared task.
Several lines of work have been conducted since the introduction of the
shared task on MMT in 2016 \cite{specia2016shared}. The majority of last year submissions
including ours \cite{caglayan2017} were based on the integration of global visual features
into various parts of the NMT architecture \cite{findings2017}. Apart from these,
hierarchical multimodal attention \cite{helcl2017} and multi-task learning \cite{elliott2017imagination}
were also explored by the participants.

This year we decided to revisit the multimodal attention \cite{caglayan2016multiatt} since
our previous observations about qualitative analysis of the visual attention was not satisfying.
In order to improve the multimodal attention both qualitatively and quantitatively, we experiment
with several refinements to it: first, we try to use different input image sizes prior to feature extraction and
second we normalize the final convolutional feature maps to assess its impact on the final MMT performance.
In terms of architecture, we propose to refine the visual features by learning an encoder-guided early spatial attention.
In overall, we find that normalizing feature maps is crucial for the multimodal attention
to obtain a comparable performance to monomodal NMT while the impact of the input image size remains unclear.
Finally, with the help of the refined attention, we obtain modest improvements in terms of BLEU \cite{Papineni:2002:acl} and METEOR \cite{Lavie:2007:acl}.

The paper is organized as follows: data preprocessing, model details and training hyperparameters are detailed
respectively in section~\ref{sec:data} and section~\ref{sec:models}.
The results based on automatic evaluation metrics are reported in section~\ref{sec:results}.
Finally the paper ends with a conclusion in section~\ref{sec:conclusion}.

\begin{figure*}[htbp]
\centering
  \includegraphics[width=\textwidth]{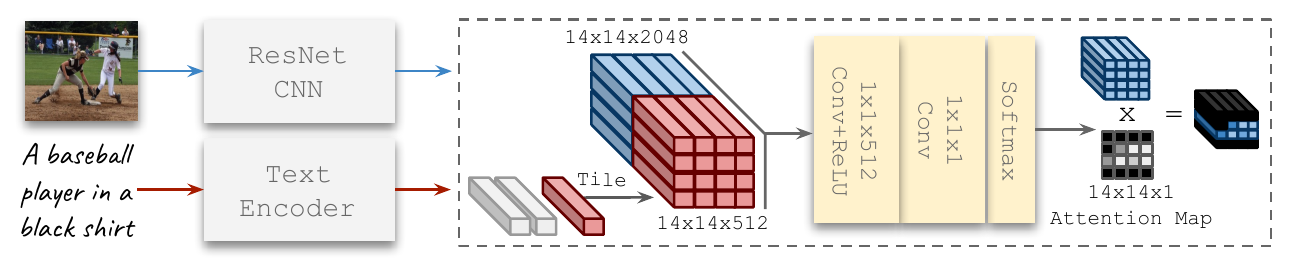}
  \caption{Filtered attention (FA): the convolutional feature maps are dynamically masked using an attention conditioned on the source sentence representation.}
  \label{fig:arch1}
\end{figure*}

\section{Data}
\label{sec:data}
We use Multi30k \cite{Elliott2016} dataset provided by the organizers which contains 29000, 1014, 1000 and 1000
English$\rightarrow$\{German,French\} sentence pairs respectively for \tm{train}, \tm{dev}, \tm{test2016} and \tm{test2017}.
A new training split of 30014 pairs is formed by concatenating the \tm{train} and \tm{dev} splits. Early-stopping is performed
based on METEOR computed over the \tm{test2016} set and the final model selection is done over \tm{test2017}.

Punctuation normalization, lowercasing and aggressive hyphen splitting were applied to all sentences prior to training.
A Byte Pair Encoding (BPE) model \cite{sennrich2015neural} with 10K merge operations is jointly learned on
English-German and English-French resulting in vocabularies of 5189-7090 and 5830-6608 subwords respectively.
\subsection{Visual Features}

Since Multi30k images involve much more complex region-level relationships and scene compositions compared to
ImageNet \cite{ILSVRC15} object classification task, we explore different input image sizes to quantify
its impact in the context of MMT since rescaling the input image has a direct effect on the size of the receptive fields of the CNN.
After normalizing the images using ImageNet mean and standard deviation, we resize and crop the images to 224x224 and 448x448.
Features are then extracted from the final convolutional layer (\tm{res5c\_relu}) of a pretrained ResNet50 \cite{he2016resnet} CNN.\footnote{We use torchvision for feature extraction.}
This led to feature maps $\mathrm{V} \in \mathbb{R}^{2048\times w\times w}$ where the spatial dimensionality $w$ is 7 or 14.

\subsubsection {Feature Normalization}
We conjecture that transferring ReLU features from a CNN into a model
that only makes use of bounded non-linearities like $sigmoid$ and $tanh$, can saturate the non-linear neurons
in the very early stages of training if their weights are not carefully initialized.
Instead of tuning the initialization, we experiment with L$_2$ normalization over the channel dimension
so that each feature vector ($\in \mathbb{R}^{2048}$) has an L$_2$ norm of 1.

\section{Models}
\label{sec:models}
In this section we will describe our baseline NMT and multimodal NMT systems.
All models use 128 dimensional embeddings and GRU \cite{cho2014gru} layers with 256 hidden states. Dropout \cite{srivastava2014dropout} is applied over source embeddings $x_s$, encoder states $\mathrm{H}^{enc}$ and pre-softmax activations $o_t$. We also apply L$_2$ regularization with a factor of $1e\mathrm{-}5$
on all parameters except biases. The parameters are initialized using the method proposed by \newcite{He_2015_ICCV} and optimized with Adam \cite{kingma2014adam}. The total gradient norm is clipped to 1 \cite{pascanu2013difficulty}. We use batches of size 64 and an initial learning rate of $4e\mathrm{-}4$.
All systems are implemented using the PyTorch version of \ti{nmtpy}\footnote{\url{github.com/lium-lst/nmtpytorch}} \cite{nmtpy}.

\subsection{Baseline NMT}
Let us denote the length of the source sentence $\{x_1, \dotsc , x_S\}$ and the target sentence $\{y_1, \dotsc , y_T\}$ by $S$ and $T$ respectively.
The source sentence is first encoded with a 2-layer bidirectional GRU to obtain the set of hidden states:
\begin{align*}
  \mathrm{H}^{enc} \leftarrow Enc(\{x_1, \dotsc , x_S\}), \mathrm{H}^{enc} \in \mathbb{R}^{S\times 512}
\end{align*}

The decoder is a 2-layer conditional GRU (CGRU) \cite{nematus} with tied embeddings \cite{press2016using}.
CGRU is a stacked 2-layer recurrence block with the attention mechanism in the middle. We use feed-forward attention \cite{Bahdanau2014} which encapsulates a learnable layer. The first decoder (which is initialized with a zero vector) receives the previous target embeddings as inputs (equation~\ref{eq:dec1}). At each timestep of the decoding stage, the attention mechanisms produces a context vector $c^{txt}_{t}$ (equation~\ref{eq:dec2}) that becomes the input to the second GRU (equation~\ref{eq:dec3}). Finally, the probability over the target vocabulary is conditioned over a transformation of the final hidden state $h^{dec_{2}}_t$ (equation~\ref{eq:dec4}, \ref{eq:dec5}).
\begin{align}
  h^{dec_{1}}_{t} &= \mathrm{DEC}_{1}(y_{t-1}, h^{dec_{2}}_{t-1})\label{eq:dec1}\\
  c^{txt}_{t} &= \mathrm{ATT}_{txt}(\mathrm{H}^{enc}, h^{dec_{1}}_{t})\label{eq:dec2}\\
  h^{dec_{2}}_{t} &= \mathrm{DEC}_{2}(c^{txt}_{t}, h^{dec_{1}}_{t})\label{eq:dec3}\\
  o_t &= \tanh (\mathbf{W_{o}} h^{dec_{2}}_t + b_{o})\label{eq:dec4}\\
  P(y_t) &= softmax(\mathbf{W_{v}} o_t)\label{eq:dec5}
\end{align}

\subsection{Multimodal Attention (MA)}
Our baseline multimodal attention (MA) system \cite{caglayan2016multiatt} applies a spatial attention mechanism \cite{xu2015show} over the visual features.
At each timestep $t$ of the decoding stage, a multimodal context vector $c_t$ is computed and given as input to the second decoder (equation ~\ref{eq:dec3}):
\begin{align}
  c_t &= \mathbf{W_{f}}\big[c^{txt}_t; \mathbf{W_{vis}}c^{vis}_t\big]\\
  c^{vis}_{t} &= \mathrm{ATT}_{vis}(\mathrm{V}, h^{dec_{1}}_{t})\label{eq:dec6}
\end{align}
Previous analysis showed that the attention over the visual features is inconsistent and weak.
We argue that this is because of the diluted relevant visual information, and the competition with the far more relevant source text information.

\subsection{Filtered Attention (FA)}
\label{sec:vda}
In order to enhance the visual attention, we propose an extension to the multimodal attention where the objective is to filter
the convolutional feature maps using the last hidden state of the source language encoder (Figure~\ref{fig:arch1}).
We conjecture that a learnable masking operation over the convolutional feature maps can help the decoder-side visual attention
mechanism by filtering out regions irrelevant to translation and focus on the most important part of the visual input.
The filtered convolutional feature map $\widetilde{\mathrm{V}}$
is computed as follows:
\begin{align}
  \mathbf{\beta}^{pre} &= ConvAtt(\big[Tile(h^{enc}_S); V\big])\\
  \widetilde{\mathrm{V}} &= \mathbf{\beta}^{pre} \odot \mathrm{V} , \mathbf{\beta}^{pre} \in \mathrm{R}^{1\times w\times w}
\end{align}

$ConvAtt$ block is inspired from previous works in visual question answering (VQA) \cite{san2015,kazemi_vqa}. It basically
computes a spatial attention distribution $\beta^{pre}$ which we further use to mask the actual convolutional features $\mathrm{V}$.
The filtered $\widetilde{\mathrm{V}}$ replaces $\mathrm{V}$ in the equation~\ref{eq:dec6} instead of being pooled
into a single visual embedding in contrast to VQA models.

\begin{table}[h]
\centering
\resizebox{.4\textwidth}{!}{%
\begin{tabular}{@{}lcc@{}}
\toprule
EN\ra DE \tm{test2017}    & BLEU 						 & METEOR         \\ \midrule
Baseline NMT      		    & 31.0 $\pm$ 0.3   & 52.1 $\pm$ 0.4 \\ \midrule
MA$_{448}$          		  & 28.6 $\pm$ 0.8   & 50.1 $\pm$ 0.3 \\
MA$_{448}$ + L$_2$-norm   & 30.8 $\pm$ 0.5   & 52.0 $\pm$ 0.2 \\ \bottomrule
\end{tabular}}
\caption{\label{tbl:l2} Impact of L$_2$ normalization on the performance of multimodal attention.}
\end{table}

\section{Results}
\label{sec:results}
We train each model 4 times using different seeds and report mean and standard deviation for the final results
using \ti{multeval} \cite{clark2011better}

\paragraph{Feature Normalization}
We can see from Table~\ref{tbl:l2} that without L$_2$ normalization, multimodal attention
is not able to reach the performance of baseline NMT. Applying the normalization consistently
improves the results for all input sizes by around $\sim$2 points in BLEU and METEOR.
From now on, we only present systems trained with normalized features.

\begin{table}[h]
\centering
\resizebox{.4\textwidth}{!}{%
\begin{tabular}{@{}lcc@{}}
\toprule
EN\ra DE \tm{test2017}        & BLEU 				     & METEOR         \\ \midrule
MA$_{224}$                    & 30.6 $\pm$ 0.4   & 51.8 $\pm$ 0.2 \\
MA$_{448}$                    & 30.8 $\pm$ 0.5   & 52.0 $\pm$ 0.2 \\ \midrule
FA$_{224}$                    & 31.5 $\pm$ 0.5   & 52.2 $\pm$ 0.5 \\
FA$_{448}$                    & 31.6 $\pm$ 0.5   & 52.5 $\pm$ 0.4 \\ \bottomrule
\end{tabular}}
\caption{\label{tbl:imsize} Impact of input image width on the performance of multimodal attention variants.}
\end{table}

\paragraph{Image Size}
Although the impact of doubling the image width and height at the input seems marginal (Table~\ref{tbl:imsize}),
we switch to 448x448 images to benefit from the slight gains which are consistent across both attention variants.

\begin{table*}[!htbp]
\centering
\renewcommand\arraystretch{1.1}
\resizebox{.75\textwidth}{!}{%
\begin{tabular}{llccc}
\toprule
\MR{2}{*}{\rm{English$\rightarrow$German}} & \MR{2}{*}{\# Params} & \MC{3}{c}{\tm{test2017} ($\mu\pm\sigma$)} \\
  & & \MC{1}{c}{\rm{BLEU}} & \MC{1}{c}{\rm{METEOR}} & \MC{1}{c}{\rm{TER}} \\ \midrule
Baseline NMT                   & 4.6M  & 31.0 $\pm$ 0.3    & 52.1 $\pm$ 0.4  & 51.2 $\pm$ 0.5  \\ \midrule
Multimodal Attention (MA)      & 10.0M & 30.8 $\pm$ 0.5    & 52.0 $\pm$ 0.2  & 51.1 $\pm$ 0.7  \\
Filtered Attention (FA)        & 11.3M & \textbf{31.6 $\pm$ 0.5}    & \textbf{52.5 $\pm$ 0.4}  & \textbf{50.5 $\pm$ 0.5}  \\
\bottomrule
\end{tabular}}
\caption{EN$\rightarrow$DE results: Filtered attention is statistically different than the NMT ($p \le 0.02$).}
\label{tbl:ende_flickr}
\end{table*}

\begin{table*}[!htbp]
\centering
\renewcommand\arraystretch{1.1}
\resizebox{.75\textwidth}{!}{%
\begin{tabular}{llccc}
\toprule
\MR{2}{*}{\rm{English$\rightarrow$French}} & \MR{2}{*}{\# Params} & \MC{3}{c}{\tm{test2017} ($\mu\pm\sigma$)} \\
  & & \MC{1}{c}{\rm{BLEU}} & \MC{1}{c}{\rm{METEOR}} & \MC{1}{c}{\rm{TER}} \\ \midrule
Baseline NMT               & 4.6M  & 53.1 $\pm$ 0.3    & 69.9 $\pm$ 0.2  & 31.9 $\pm$ 0.8  \\ \midrule
Multimodal Attention (MA)  & 10.0M & 52.6 $\pm$ 0.3    & 69.6 $\pm$ 0.3  & 31.9 $\pm$ 0.4  \\
Filtered Attention (FA)    & 11.3M & 52.8 $\pm$ 0.2    & 69.6 $\pm$ 0.1  & 31.9 $\pm$ 0.1  \\
\bottomrule
\end{tabular}}
\caption{EN$\rightarrow$FR results: multimodal systems are not able to improve over NMT in terms of automatic metrics.}
\label{tbl:enfr_flickr}
\end{table*}

\begin{table}[]
\centering
\resizebox{.4\textwidth}{!}{%
\begin{tabular}{@{}lccc@{}}
\toprule
EN\ra DE        & BLEU & METEOR & TER  \\ \midrule
MeMAD$\dagger$  & 38.5 & 56.6  & 44.5 \\
UMONS$\star$   	& 31.1 & 51.6  & 53.4 \\
LIUMCVC-FA$\star$  & 31.4 & 51.4  & 52.1 \\
LIUMCVC-NMT$\star$ & 31.1 & 51.5 & 52.6 \\ \midrule\midrule
EN\ra FR 				& 		 & 			 & 			\\ \midrule
CUNI$\dagger$   & 40.4 & 60.7  & 40.7 \\
LIUMCVC-FA$\star$ 	& 39.5 & 59.9  & 41.7 \\
LIUMCVC-NMT$\star$ & 39.1 & 59.8 & 41.9 \\
\bottomrule
\end{tabular}}
  \caption{Official \tm{test2018} results ($\dagger$: Unconstrained, $\star$: Constrained.)}
\label{tbl:res_2018}
\end{table}

\subsection{Monomodal vs Multimodal Comparison}
We first present the mean and standard deviation of BLEU and METEOR over 4 runs on the internal test set \tm{test2017} (Table~\ref{tbl:ende_flickr}).
With the help of L$_2$ normalization, MA system almost reaches the monomodal system but fails to improve over it.
On the contrary, the filtered attention (FA) mechanism improves over the baseline and produces hypotheses
that are statistically different than the baseline with $p \le 0.02$.

The improvements obtained for EN\ra DE language pair are not reflected on the EN\ra FR performance.
One should note that the hyperparameters from EN\ra DE task were transferred to EN\ra FR without
any other tuning.

The automatic evaluation of our final submissions (which are ensembles of 4 runs) on the official test set \tm{test2018} is
presented in Table~\ref{tbl:res_2018}. In addition to our submissions, we also provide the best constrained and unconstrained
systems\footnote{\url{www.statmt.org/wmt18/multimodal-task.html}} in terms of METEOR. However, it should be noted that the submitted systems will be primarily evaluated using
human direct assessment.

On EN\ra DE, our constrained FA system is comparable to the constrained UMONS submission. On EN\ra FR,
our submission obtained the highest automatic evaluation scores among the constrained submissions
and is slightly worse than the unconstrained CUNI system.

\section{Conclusion}
\label{sec:conclusion}
MMT task consists of translating a source sentence into a target language with the help of an image representing the source sentence.
The different level of relevance of both input modalities makes it a difficult task where the image should be used with parsimony.
With the aim of improving the attention over visual input, we introduced a filtering technique to allow the network to ignore irrelevant parts of the image that should not be considered during decoding. This is done by using an attention-like mechanism between the source sentence and the convolutional feature maps. Results show that this mechanism significantly improves the results for English\ra German on one of the test sets.
In the future, we plan to qualitatively analyze the spatial attention and try to improve it further.

\section*{Acknowledgments}
This work was supported by the French National Research Agency (ANR) through the CHIST-ERA M2CR project\footnote{\url{http://m2cr.univ-lemans.fr}}, under the contract number ANR-15-CHR2-0006-01 and by MINECO through APCIN 2015 under the contract number PCIN-2015-251.

\bibliography{paper}
\bibliographystyle{acl_natbib_nourl}

\end{document}